\documentclass[preprint,12pt]{elsarticle}

\usepackage{amsmath}

\usepackage[dvipsnames]{xcolor}
\usepackage{amssymb}
\usepackage{tabularx}
\biboptions{sort&compress}

\journal{Mechanical Systems and Signal Processing}

\begin{document}

\begin{frontmatter}



\title{Data-driven Nonlinear Modal Analysis with Physics-constrained Deep Learning: Numerical and Experimental Study}

\author[1]{Abdolvahhab Rostamijavanani\corref{cor1}}
\cortext[cor1]{Corresponding author}
\ead{arostami@mtu.edu}
\author[1]{Shanwu Li}
\author[1]{Yongchao Yang}

\affiliation[1]{organization={Department of Mechanical Engineering - Engineering Mechanics, Michigan Technological University},
            addressline={1400 Townsend Drive}, 
            city={Houghton},
            postcode={49931-1295}, 
            state={Michigan},
            country={USA}}

\begin{abstract}
To fully understand, analyze, and determine the behavior of dynamical systems, it is crucial to identify their intrinsic modal coordinates. In nonlinear dynamical systems, this task is challenging as the modal transformation based on the superposition principle that works well for linear systems is no longer applicable. To understand the nonlinear dynamics of a system, one of the main approaches is to use the framework of Nonlinear Normal Modes (NNMs) which attempts to provide an in-depth representation. In this research, we examine the effectiveness of NNMs in characterizing nonlinear dynamical systems. Given the difficulty of obtaining closed-form models or equations for these real-world systems, we present a data-driven framework that combines physics and deep learning to \textit{identify} the nonlinear modal transformation function of NNMs from response data only. We assess the framework's ability to represent the system by analyzing its mode decomposition, reconstruction, and prediction accuracy using a nonlinear beam as an example. Initially, we perform numerical simulations on a nonlinear beam at different energy levels in both linear and nonlinear scenarios. Afterward, using experimental vibration data of a nonlinear beam, we isolate the first two NNMs.  \textcolor{black}{It is observed that the NNMs' frequency values increase as the excitation level of energy increases, and the configuration plots become more twisted (more nonlinear). In the experiment, the framework successfully decomposed the first two NNMs of the nonlinear beam using experimental free vibration data and captured the dynamics of the structure via prediction and reconstruction of some physical points of the beam. }
\end{abstract}




\end{frontmatter}


\section{Introduction}
\noindent It is imperative to identify the appropriate modal coordinates of dynamical systems in order to analyze, and characterize the dynamics that underlie these systems for a number of purposes including system identification~\cite{brunton2016discovering}, modal analysis~\cite{taira2017modal}, control~\cite{lu2013nonlinear}, and reduced-order modeling of a wide variety of dynamical systems~\cite{haller2016nonlinear,komatsu1991experimental,likins1967modal,touze2008reduced,bladh2002dynamic,bladh2001component,peeters2009modern,vakakis1997non,kerschen2009nonlinear,li2021hierarchical}. In linear dynamical systems, linear normal and eigenmodes (LNMs) are represented universally through modal transformations~\cite{heylen1997modal}, which provide a detailed description of the dynamical characteristics underlying them. \textcolor{black}{Proper orthogonal mode decomposition~(POD)~\cite{noor1980reduced,kerschen2000proper,mortara2000proper,dowell1999reduced} and dynamic mode decomposition (DMD)~\cite{schmid2010dynamic,kutz2016dynamic,tu2013dynamic,schmid2011applications} are data-driven techniques used to extract coherent structures and dynamics from dynamical systems }. However, the applicability of modal superposition is limited to linear systems and there exist no such general mathematical framework for representing nonlinear dynamical systems. Simply using LNMs with these linear methods typically causes significant errors in modeling nonlinear dynamics, especially in strongly nonlinear cases. Seeking some nonlinear generalization of the modal superposition is thus widely studied for more accurate representation and characterization of nonlinear dynamical systems~\cite{mezic2005spectral,mezic2013analysis,arbabi2017ergodic,kerschen2009nonlinear,shaw1991non,shaw1993normal,rand1971higher,manevich1972periodic,vakakis1991analysis}. As a pioneering work, nonlinear normal modes (NNMs)~\cite{rosenberg1960normal}, originally introduced by Rosenberg and extensively studied by numerous researchers~\cite{kerschen2009nonlinear,shaw1991non,shaw1993normal,rand1971higher,manevich1972periodic,vakakis1991analysis}, extend linear normal modes (LNMs) in order to capture nonlinear dynamical systems' intrinsic invariance properties. \textcolor{black}{According Rosenberg's NNMs, expanding its application to non-conservative systems is not a straightforward task}~\cite{kerschen2009nonlinear,renson2016robust,peeters2011modal,touze2006nonlinear}. By introducing an invariant manifold definition of NNMs, Shaw and Pierre~\cite{shaw1991non,shaw1993normal} addressed this limitation by extending the invariance of linear manifolds to nonlinear manifolds~\cite{cirillo2016spectral,worden2017machine}.

In the real world, dynamical systems are generally unknown without the knowledge of closed-form models or equations, and we often only have limited measurements available. Recent advances in deep learning, including the Physics Informed Neural Networks (PINNs)~\cite{raissi2019physics,raissi2020hidden} framework, have facilitated the development of data-driven methods that enable nonlinear analysis of dynamical systems such as Koopman operator, which provides a linear representation of nonlinear dynamics~\cite{lusch2018deep,yeung2019learning,takeishi2017learning,otto2019linearly}. It is noteworthy that a low-dimensional autoencoder in the form of a deep neural network (DNN) was presented to capture Koopman modal coordinates for a comprehensive representation of continuous-spectrum nonlinear dynamical systems~\cite{lusch2018deep}. Furthermore, a Gaussian-based machine learning approach~\cite{worden2017machine,dervilis2019nonlinear} was presented for the identification of NNMs, and a nonlinear manifold study using pattern recognition, and data-driven identification of NNMs using a physics-informed DNN~\cite{shanwu}. While most dynamical systems studied by researchers with deep/machine learning techniques are numerically simulated systems with ideal data, we aim to use deep learning techniques for system identification and modal analysis of a laboratory nonlinear beam with non-ideal (e.g., partial and noisy) measurements. 

During numerical simulations, we have a greater degree of flexibility in determining design parameters. For example, we can change a system's stiffness or nonlinearity level numerically, whereas in experiments we have fewer options to change the parameters of the structure or it is more difficult to have these changes made. In addition, the data acquired by data acquisition systems are not clean and require pre-processing steps such as data filtering and cleaning. Furthermore, we are not able to control the complexity of the structure easily, so an experimental system may exhibit chaotic behavior. This combination of challenges makes any data-driven model for modal analysis and system identification difficult to use in experimental systems. Therefore, we present a data-driven approach adapted to NNM constraints that capture the dynamics of a nonlinear system using only measurement data.

\section{Problem formulation}\label{sec:2}
As stated by the general equation of motion, the free response of a system with $\mathrm{N}$ degrees of freedom (N-DOF) is taken into consideration as:
\begin{equation}\label{eq1}
{\mathrm{M\ddot x}} + {\mathrm{C\dot x}} + {\mathrm{Kx}} + {\mathrm{g}}\left( {{\mathrm{\ddot x}},{\mathrm{\dot x}},{\mathrm{x}}} \right) = {\mathrm{f}\left(t\right)}
\end{equation}
\noindent In this equation, the mass, damping, and stiffness matrices (denoted as $\mathrm{M}$, $\mathrm{C}$, and $\mathrm{K}$ respectively) are used to describe the dynamics of the system. The displacement vector ($\bold x$) is represented in n-dimensional space ($\mathrm{x\in R^n}$) and $\mathrm{g}$ represents the nonlinear term in the equation. Assuming a state space transformation:
\begin{equation}
\begin{split}
    \mathrm{\bold z=\{\bold x,\dot{\bold x}\}}\\
    \mathrm{\bold z^{t+1}=F\left(\bold z^{t}\right)}
    \end{split}
\end{equation}
where $\mathrm{z \in R^{2n}}$ is a vector in the state space that is measured by sensors or computed numerically, and $\mathrm{F}$ is the function that defines the dynamics of the system and maps the current state to the future state. \par
\textcolor{black}{Nonlinear modal transformation can be achieved in both forward and inverse directions by~(see Fig. \ref{fig:architecture}):}
\begin{equation}
\begin{split}
    \textcolor{black}{\varphi^{t}=\vartheta(z^{t})}
    \\
    \textcolor{black}{z^{t}=\vartheta^{-1}(\varphi^{t})}
   \end{split} 
\end{equation}
\textcolor{black}{where \textit{$\varphi$} denotes modal coordinates and $\vartheta$ represents the modal transformation function.}

Through nonlinear transformations of the intrinsic modal coordinates, NNMs can represent the nonlinear dynamics of a system in the invariant modal space (manifolds). As an extension of LNMs, NNMs are able to represent nonlinear systems with the same number of modal coordinates as the original coordinates:
\begin{equation}
    \mathrm{\varphi^{t+1}=G\left(\varphi^{t}\right)}
\end{equation}
\textcolor{black}{where function $\mathrm{G}$ is used to represent the modal state transition.} It is important to note that Nonlinear Normal Modes (NNMs) use nonlinear transformations of intrinsic modal coordinates to represent nonlinear dynamics. These coordinates include displacement and velocity fields, as shown in Fig. \ref{fig:architecture}. The NNM-associated physics constraints are integrated into our data-driven modal-analysis-based framework to identify the nonlinear modal transformation function \textcolor{black}{$\vartheta$} and its generalized inverse \textcolor{black}{$\vartheta^{-1}$}, and the modal dynamics function $G$ from the response data $\mathrm{z}$ only. Then, these identified nonlinear functions are required to perform data-driven modal analysis for the measured system without the governing equation, including the decomposition of the original response $\mathrm{z}$ into nonlinear modal response $\varphi$, extraction/identification of the NNMs with invariant manifolds from the original response data $\mathrm{z}$, and future state prediction in both modal and original space.

\section{Physics-constrained deep autoencoder framework}
\textcolor{black}{This work aims to evaluate the capability of a data-driven nonlinear modal analysis framework with nonlinear normal modes-embedded deep neural network.} The performance of the modal analysis framework is assessed through mode decomposition, prediction, and reconstruction of the nonlinear beam response. The architecture of the physics-integrated deep autoencoder used is illustrated in Fig.~\ref{fig:architecture}. Autoencoders \cite{wang2016auto,lore2017llnet} use bottleneck latent spaces to capture the key features of the original data. When applying NNMs to nonlinear beams, the original coordinates are transformed to latent intrinsic coordinates (NNM modal coordinates) using an encoder block,~$\mathrm{{\vartheta }:\;{\mathbb{R}^{2n}} \to {\mathbb{R}^{2s}}}$, where $\mathrm{n}$ and $\mathrm{s}$ are the dimensions of the original coordinates and the latent coordinates, respectively. For NNMs, the number of latent coordinates is the same as the number of original coordinates, $\mathrm{s=n}$. The encoder's last layer is representative of the most prominent characteristics of the corresponding NNM modal coordinates. The input data is represented by a $\mathbb{R}\in {2n}$ vector, where $\mathrm{2n}$ refers to the state space dimensions of an $\mathrm{n}$-DOF system. With NNMs, the latent space tensor dimensions are the same as the input shape.

 A nonlinear beam is a dynamic system that can be modeled by a second-order ordinary differential equation (ODE). Thus, every pair of latent coordinates corresponds to a displacement and a velocity of \textcolor{black}{one mode} in the DNN presented for NNMs. Additionally, the number of latent dimensions is the same as the number of original dimensions, i.e., $\mathrm{s=n}$. Hence, each pair of latent coordinates is expected to correspond to a single NNM modal coordinate.

\subsection{Learning nonlinear normal modes (NNMs)}
\begin{figure}[htbp]
	\centering
\includegraphics[width=1\textwidth]{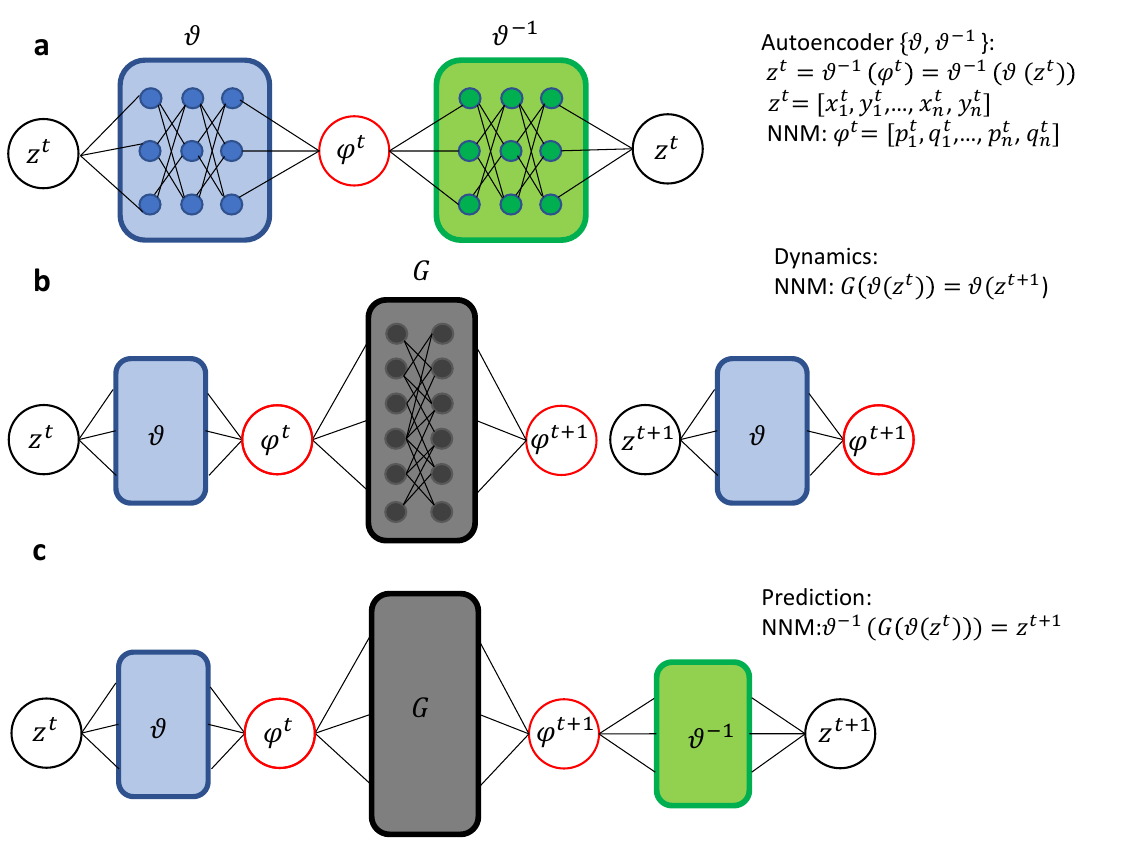}
	\caption{Our physics-constrained deep autoencoder architecture: \textbf{a} The framework includes a deep autoencoder that transforms system states $z=(x,y)$ into intrinsic coordinates $(p,q)$ or $\varphi$ through the function ${\mathrm{\varphi}} = \vartheta \left( {\mathrm{z}} \right)$. The autoencoder then decodes the intrinsic coordinates back to the original coordinates using ${\mathrm{z}} = {\vartheta ^{ - 1}}\left( {\mathrm{\varphi}} \right)$. Additional physics-based constraints can be applied to the intrinsic coordinates to convert them to desired modal coordinates. \textbf{b} A dynamics block ($G$) is also implemented, which advances intrinsic coordinates over time and ensures that encoding the next original coordinates is equivalent to advancing the current intrinsic coordinates.
\textbf{c} By combining the encoder, dynamics block, and decoder in the appropriate order, intrinsic coordinates can be determined for predicting future states.
It is important to note that the decoder is not the exact inverse of the encoder, but it is approximated as closely as possible through a reconstruction loss function.}
	\label{fig:architecture}
\end{figure}

The NNM-physics-constrained autoencoder integrates the physics of NNMs into the deep learning framework. The following is a summary of the loss function:
\begin{equation}\label{eq 333}
    \mathrm{\mathcal{L}_{NNM}= {\alpha _{rec}}{\mathcal{L}_{rec}} + {\alpha _{corr}}{\mathcal{L}_{corr}} + {\alpha _{evol}}{\mathcal{L}_{evol}} + {\alpha _{prd}}{\mathcal{L}_{prd}} + {\alpha _{vel}}{\mathcal{L}_{vel}}+{\alpha _{spar}}{\mathcal{L}_{spar}}}
\end{equation}
where the overall loss function for the NNM-AE is denoted as $\mathcal{L}_{NNM}$, and the weights of each loss function are listed in Table.\ref{tbl2}. The loss functions used include: $\mathcal{L}_{rec}$ for reconstruction in the original coordinates, $\mathcal{L}_{corr}$ for independence between modal coordinates~(latent coordinates), $\mathcal{L}_{evol}$ for dynamics in the latent space, $\mathcal{L}_{prd}$ for prediction in the original coordinates, $\mathcal{L}_{vel}$ for incorporating the state-space format into the latent space, and $\mathcal{L}_{spar}$ for sparsity of the latent space coordinates. Each of these loss functions is described in more detail below, and the mean squared error (MSE) between two matrices or vectors is denoted as $||~,~||_{MSE}$ (e.g. between the reconstructed trajectory and the original trajectory). 
\begin{enumerate}
\itemsep=3pt
\item The encoder block (shown as a blue block in Fig.~\ref{fig:architecture}) performs the forward nonlinear transformation to convert the original coordinates into latent intrinsic coordinates. The decoder block (shown as a green block) is responsible for reversing this process, converting the latent/modal coordinates back to the original coordinates. This operation is associated with a loss called reconstruction, which ensures that the autoencoder is able to accurately reconstruct the original coordinates from the latent coordinates. As a result, this loss is minimized: 
\begin{equation}
\mathcal{L}_{rec}=\frac{1}{ns}\sum_{i=1}^{i=ns}||\bold z^{t},\vartheta^{-1}\left(\vartheta\left(\bold z^{t}\right)\right)||^{(i)}_{{\text{MSE}}}
\end{equation}
 where the notation $\mathrm{ns}$ represents the number of training samples, and $\mathrm{(i)}$ denotes the index of a specific sample.
\item To ensure that the nonlinear modal coordinates are independent, we impose modal-uncorrelatedness by minimizing the loss function ${\mathcal{L}_{corr}}$, which enforces independence of the NNM modal coordinates:
\begin{equation}
\begin{split}
\frac{1}{ns}\sum_{i=1}^{i=ns}||Corr\left(\boldsymbol p\right),I_{s\times s}||^{(i)}_{{\text{MSE}}}\\
\frac{1}{ns}\sum_{i=1}^{i=ns}||Corr\left(\boldsymbol q\right),I_{s\times s}||^{(i)}_{{\text{MSE}}}\\
\frac{1}{ns}\sum_{i=1}^{i=ns}||Corr\left(\boldsymbol {\dot{p}}\right),Corr\left(\boldsymbol p\right)||^{(i)}_{{\text{MSE}}}
\end{split}
\end{equation}
where the identity matrix is represented by \textit{\textbf{I}}, and the correlation matrix is represented by \textit{\textbf{Corr}}. The displacement matrix is denoted as $\boldsymbol p=[p_{1},p_{2},...,p_{s}]$ and the velocity matrix is denoted as $\boldsymbol q=[q_{1},q_{2},...,q_{s}]$, where each $p_{i}$ or $q_{i}$ has a length of $T$. $\boldsymbol{\dot{p}}$ represents the time derivative of the displacement matrix ($\frac{\Delta \boldsymbol p}{\Delta t}$, where $\Delta t$ is provided as an input to the network) and \emph \textit{\textbf{s}} is the degree of freedom in the system. This loss function is used to enforce independence of displacement and velocity modal decomposition.
\item Identifying the evolution function by Dynamics block: Latent space evolution (Nonlinear Dynamics). In Dynamics block (grey color in Fig.~\ref{fig:architecture}), the networks determine the evolution of the system state by using the initial time response of each example of training, which is accomplished by minimizing the residual of the following expression:
\begin{equation}
 \mathcal{L}_{evol}=\frac{1}{ns}\sum_{i=1}^{i=ns}||\vartheta\left(\bold z^{t+1}\right),  G\left(\vartheta\left(\bold z^{t}\right)\right)||^{(i)}_{{\text{MSE}}} 
\end{equation}
 where $\mathrm{G}$ represents the dynamic block, which is considered to be a nonlinear embedded dynamic with nonlinear activation functions (Relu function). For $\mathrm{m}$-time-steps prediction, we minimize the loss $||\vartheta\left(\bold z^{t+m}\right)-G\left( G\left(G...\left(\vartheta\left(\bold z^{t}\right)\right)\right)\right)||_{{\text{MSE}}}$ , where state space has to pass through the nonlinear dynamics block ($\mathrm{G}$) $\mathrm{m}$ times.
\item Prediction by considering the autoencoder and dynamics block: Prediction in the original coordinate system. The decoder aims to transform the prediction of evolution in latent coordinates to the original coordinates by minimizing the following: 
\begin{equation}
 \mathcal{L}_{prd}=\frac{1}{ns}\sum_{i=1}^{i=ns}||\bold z^{t+1},  \vartheta^{-1}\left(G\left(\vartheta\left(\bold z^{t}\right)\right)\right)||^{(i)}_{{\text{MSE}}}  
\end{equation}
 or generally for $\mathrm{m}$-time-steps prediction, we minimize $\frac{1}{ns}\sum_{i=1}^{i=ns}||\bold z^{t+m}-\vartheta^{-1}\left(G\left(G\left(G...\left(\vartheta\left(\bold z^{t}\right)\right)\right)\right)\right)||^{(i)}_{{\text{MSE}}}$

\item Velocity loss.  The constraint that a latent dimension should be a combination of the displacement and velocity fields of a modal coordinate~($\mathrm{p,q}$) is implemented as a loss function as:\begin{equation}
\mathcal{L}_{vel}=\frac{1}{ns}\sum_{i=1}^{i=ns}||\frac{\Delta{p_i}}{\Delta t}, q_i||^{(i)}_{{\text{MSE}}}
\end{equation}
where $\mathrm{\Delta t}$ is the time step provided as input to the network.
\item Sparsity. The following loss function is minimized to enforce zero-mean oscillation in NNM modal coordinates:
\begin{equation}
 {\mathcal{L}_{spar}}=\frac{1}{ns}\sum_{i=1}^{i=ns}||\varphi^t, \Theta_{s\times s}||^{(i)}_{{\text{MAE}}}  
\end{equation}
where $\mathrm{\Theta_{s\times s}}$ is a zero matrix with the size of $\mathrm{s\times s}$ and $\mathrm{||~,~||_{MAE}}$ is the mean absolute error between two matrices or vectors. This loss function ensures sparsity of the observed behavior within the identified modal space, thus improving the accuracy of identifying the modal space where the dynamics of the system can be described using fewer coordinates than in the original state space.
\end{enumerate}
\subsection{Network architecture and training}
The DNN presented in this study includes three models: the Encoder, the Decoder, and the Dynamics block. Table.\ref{tbl1} provides details about the number of layers and neurons for each model. Each model, which is a type of multilayer perception model, performs the following tasks:\par 
\textit{Encoder}: The encoder model's goal is to convert the original coordinates into modal (latent) coordinates through a forward modal transformation. The output of the encoder represents the modal coordinates, which are then processed through the dynamics block and decoder. As a result, this model has the following loss functions: $\mathcal{L}_{rec}$, $\mathcal{L}_{prd}$, $\mathcal{L}_{corr}$, $\mathcal{L}_{vel}$, and $\mathcal{L}_{spar}$.\par
\textit{Decoder}: The decoder model is responsible for converting the modal coordinates from the latent space back to their original coordinates. To train this model, the reconstruction loss ($\mathcal{L}_{rec}$) and prediction loss ($\mathcal{L}_{prd}$) are taken into account.\par
\textit{Dynamics block}: This model aims to predict the dynamics of a system by mapping intrinsic modal coordinates to a set of specific future time steps. To achieve this, two loss functions need to be trained as part of the overall framework: $\mathcal{L}_{evol}$ and $\mathcal{L}_{prd}$.

Note that each of the three models is trained simultaneously, in other words, all weights are shared and modified at the same time during the training process. The Xavier initialization method~\cite{glorot2010understanding} was used to establish each model's initial weights. The hidden layers are defined as ${\mathrm{Wa + b}}$ followed by a nonlinear activation where $\mathrm{W}$ and $\mathrm{b}$ are weights and biases respectively and $\mathrm{a}$ corresponds to input data. The Xavier initialization method is designed to produce a random number that is uniformly distributed along a range of $\mathrm{-\frac{1}{\sqrt{\eta}}}$ and $\mathrm{\frac{1}{\sqrt{\eta}}}$, where $\eta$ represents the number of inputs to each node. Throughout a variety of training sessions, we analyze the performance of the DNN (hyperparameters-tuning).  We examined different sets of hyperparameters (weights of loss functions) and our findings are based on those hyperparameters associated with the lowest validation error (Table.~\ref{tbl2}). We use nonlinear activation functions for NNMs' DNNs because we seek nonlinear modal transformations as well as nonlinear mapping for the dynamics of the system. The Encoder, Decoder, and Dynamics blocks of NNMs DNN use Relu (for fast training runs) activation function, which can be expressed as: $\mathrm{f(\zeta)=max(0,\zeta)}$
where $\mathrm{\zeta=Wa+b}$. Adam optimizer\cite{kingma2014adam} with a small learning rate, $\alpha_{opt}=0.0001$ is used for both DNNs.

\begin{table}
\caption{Network Architecture. Note: there are the same number of neurons in each layer for each block}\label{tbl1}
\begin{tabular}{lp{3.5cm}p{3.5cm}p{3.5cm}}
		\hline\noalign{\smallskip}
Block  & Layer type & Number of Layers  & Number of Neurons Per Layer\\
\noalign{\smallskip}\hline\noalign{\smallskip}
Encoder & Dense & 3 & 128 \\
Dynamic & Dense & 4 & 256 \\
Decoder & Dense & 3 & 128 
 \\
\noalign{\smallskip}\hline
\end{tabular}
\end{table}

\begin{table}
\caption{Weights of loss functions for each DNN.}\label{tbl2}
\begin{tabular}{lp{1.5cm}p{1.5cm}p{1.5cm}p{1.5cm}p{1.5cm}p{1.5cm}p{1.5cm}}
\hline\noalign{\smallskip}
$~$ & $\alpha_{rec}$  & $\alpha_{evol}$ & $\alpha_{prd}$  & $\alpha_{corr}$ & $\alpha_{vel}$ & $\alpha_{spar}$\\
\noalign{\smallskip}\hline\noalign{\smallskip}
NNM & 1 & 1000 & 1000 & 1 & 1 & 1 \\
\noalign{\smallskip}\hline
\end{tabular}
\end{table}
\begin{figure}[htbp]
	\centering
\includegraphics[width=1\textwidth]{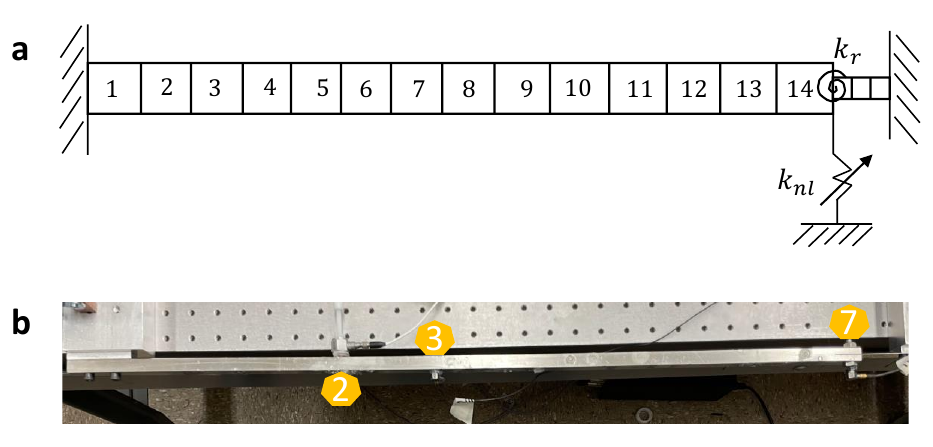}
	\caption{Nonlinear beam: \textbf{a} The finite element model of a nonlinear beam. Nonlinear spring is shown as $k_nl$ and rotational spring is shown in the junction of main and thin beam as $k_r$. \textbf{b} The experimental nonlinear beam. Shaker is located at position 2 and we measure the acceleration of points 3 and 7.}
	\label{beam}
\end{figure}
\section{Results}
\textcolor{black}{In this section, we have two goals: (i) Firstly, we present the procedure for NNMs mode isolation of a \textit{numerical} nonlinear beam. Using the so-called force appropriation approach, we demonstrate the procedure for extracting the NNMs of the structure. Additionally, we investigate the effect of nonlinearity and energy level on the NNMs of the beam. (ii) In the next step, we present a data-driven approach for NNMs mode decomposition of an \textit{experimental} nonlinear beam using a physics-constraint deep neural network. It should be noted that, although we present a data-driven method for experimental data only, the numerical simulation data are also used in the training process of the deep neural network.}

\subsection{Numerical nonlinear beam}
The numerical simulation of a nonlinear beam is taken as an example to demonstrate the presented data-driven nonlinear modal analysis method. The mathematical model of the nonlinear beam is obtained using the finite element method. The equations of motion are then determined from this model:
\begin{equation}\label{eq1}
{\mathbf{M\ddot x}} + {\mathbf{C\dot x}} + {\mathbf{Kx}} + {\mathbf{f_{nl}}} = \mathbf{f(t)}
\end{equation}
\noindent where $\mathbf{M}$, $\mathbf{C}$, $\mathbf{K}$, $\mathbf{f_{nl}}$, and $\mathbf{f(t)}$  are mass, damping, stiffness matrices, nonlinear force, and external force respectively. The simulation of the main and thin beams is accomplished by using 14 Euler-Bernoulli beam elements and 3 Euler-Bernoulli beam elements respectively. A linear rotational stiffness ($k_r$, as shown in Fig.\ref{beam}) is employed to model the junction between the two beams. A grounded cubic spring is utilized to simulate the nonlinear behavior of the thin beam at the intersection of the main and thin beams:
\begin{equation}
    \mathbf{f_{nl}(x)=k_{nl}|x^3|sign(x)}
\end{equation}
The geometrical stiffening effect of the thin beam is included in the simulation by using a cubic term. The nonlinear coefficient, $\mathbf{k_{nl}}$, is set to $15e9$. Linear proportional damping (Rayleigh damping) is applied to model the dissipated forces in the structure. The damping matrix, $\mathbf{C}$, is represented as follows:
\begin{equation}
    \mathbf{C =3e-7K + 5M}
\end{equation}
\indent The nonlinear beam has fixed-fixed boundary conditions at its left and right sides, which constrain its displacements and rotations.

\subsubsection{Numerical: Linear Analysis}
\begin{figure}[htbp]
	\centering
\includegraphics[width=1\textwidth]{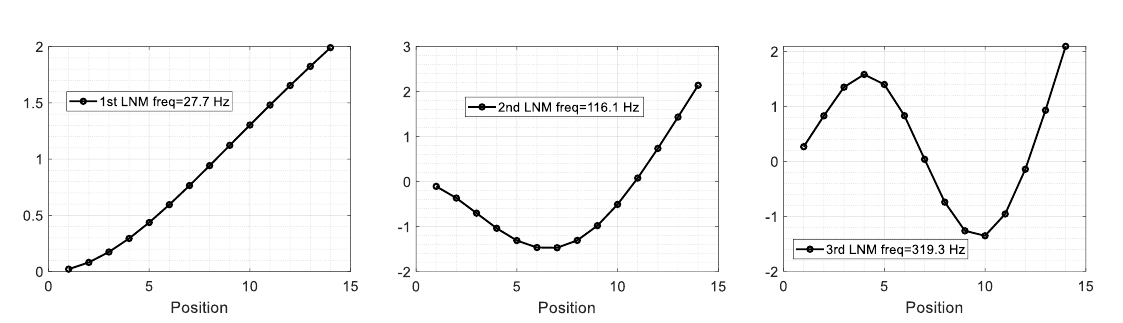}
	\caption{First three natural frequencies and corresponding mode shapes of the nonlinear beam}
	\label{fig:natural}
\end{figure}

\textcolor{black}{To ensure the accuracy of the nonlinear analysis of the nonlinear beam, we first calculate its natural frequencies and corresponding mode shapes. This is crucial as the natural frequencies provide important information necessary for the nonlinear analysis.} Therefore, for linear modal analysis, we use the eigenvalue approach to extract the natural frequencies and mode shapes. Fig. \ref{fig:natural} illustrates the first three natural frequencies and the corresponding mode shapes.
\subsubsection{Numerical: Force appropriation}
The force appropriation method for extracting a single NNM is briefly explained here. The structure is excited at various frequencies at a specific energy level and the displacement magnitudes are measured at a specific point on the structure. These measurements are plotted as two vectors, one for frequencies and the other for displacement magnitudes. The frequency at which there is a significant change in displacement corresponds to the NNM at that specific energy level (since NNM is energy-dependent, different energy levels will have different NNM values). In numerical analysis, a step-sin force can be applied as the external force in equation. \eqref{eq1} to implement the force appropriation method and isolate a single NNM at a specific energy level using fixed amplitude step-sin excitation: $\mathbf{f(t)= A~sin(\omega t)}$). \textcolor{black}{In order to examine the impact of the structural nonlinearity intensity on the NNMs, we present two separate case studies in the following subsections: one with weakly nonlinear features and another with highly nonlinear features.}
\subsubsection{Numerical: Weakly nonlinear structure}
\begin{figure}[htbp]
	\centering
\includegraphics[width=1\textwidth]{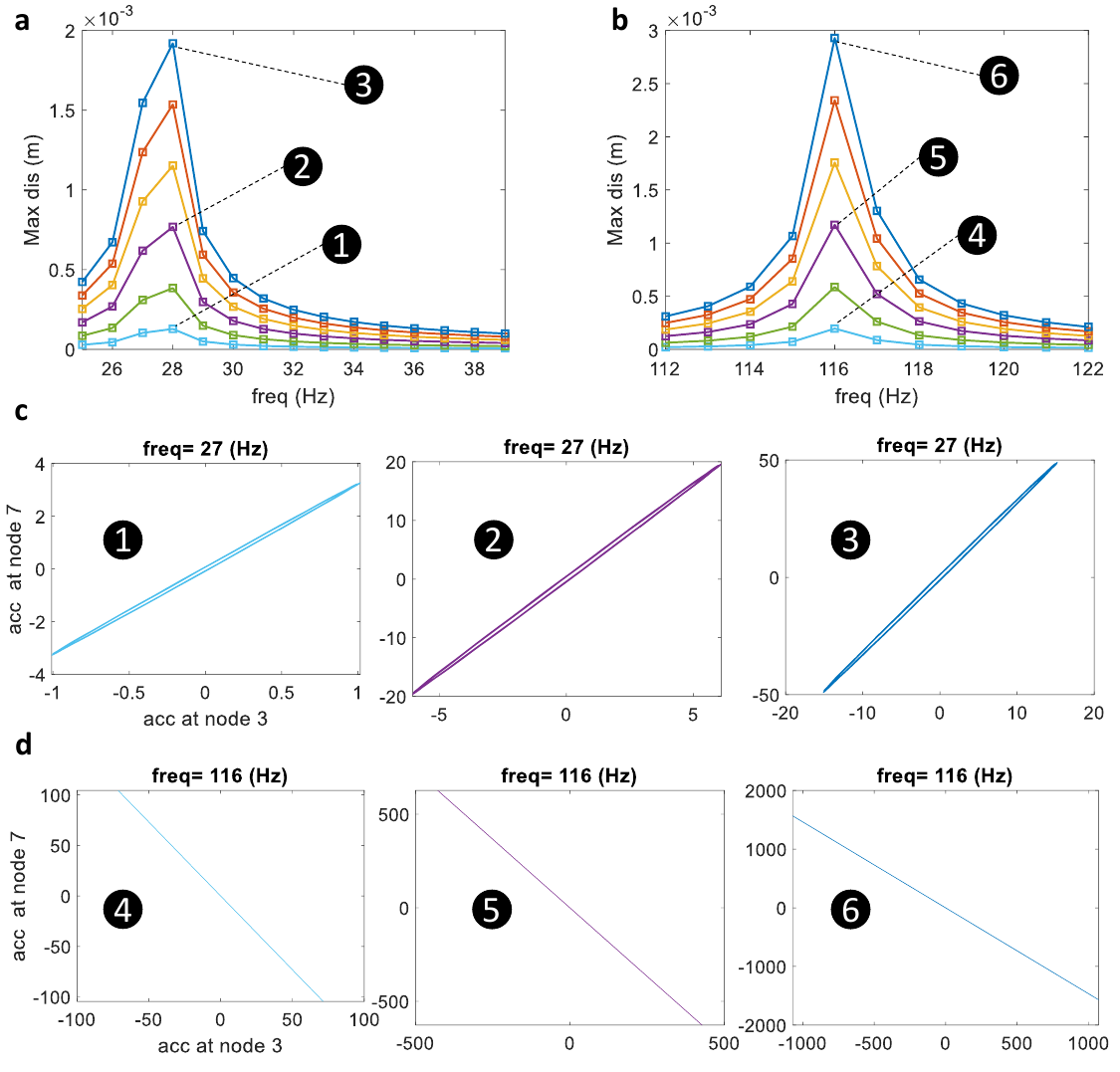}
	\caption{Nonlinear mode  isolation of first NNM of a weakly nonlinear beam. \textbf{a} First NNM for different energy levels. \textbf{b} Second NNM for different energy levels. \textbf{c} Configuration plots for three different energy levels (marked in plot \textbf{a}). \textbf{d} Configuration plots of three different frequencies at a specific level of energy (as marked in plot \textbf{b})}
	\label{fig:num_linear}
\end{figure}

 The intensity of nonlinearity in the structure can be determined by the coefficient $k_{nl}$~(as shown in Fig. \ref{beam}). A weakly nonlinear beam is modeled by setting $k_{nl}=0.1$, which results in a nearly linear case study. The goal is to extract the first two NNMs of the structure. To do this, the step-sin force appropriation method is applied at different energy levels around each natural frequency. It is clear that the isolated NNMs (1st and 2nd NNMs) remain unchanged at different energy levels. This is because the structure is not nonlinear and therefore not dependent on energy. Additionally, the configuration plots for accelerometers at positions~(nodes) 3 and 7 for the first and second modes are illustrated in Fig. \ref{fig:num_linear} (c) and Fig. \ref{fig:num_linear} (d), respectively. As observed, the configurations are flat lines, indicating that the structure is linear.
\subsubsection{Numerical: Highly nonlinear structure: first NNM isolations}
\begin{figure}[htbp]
	\centering
\includegraphics[width=1\textwidth]{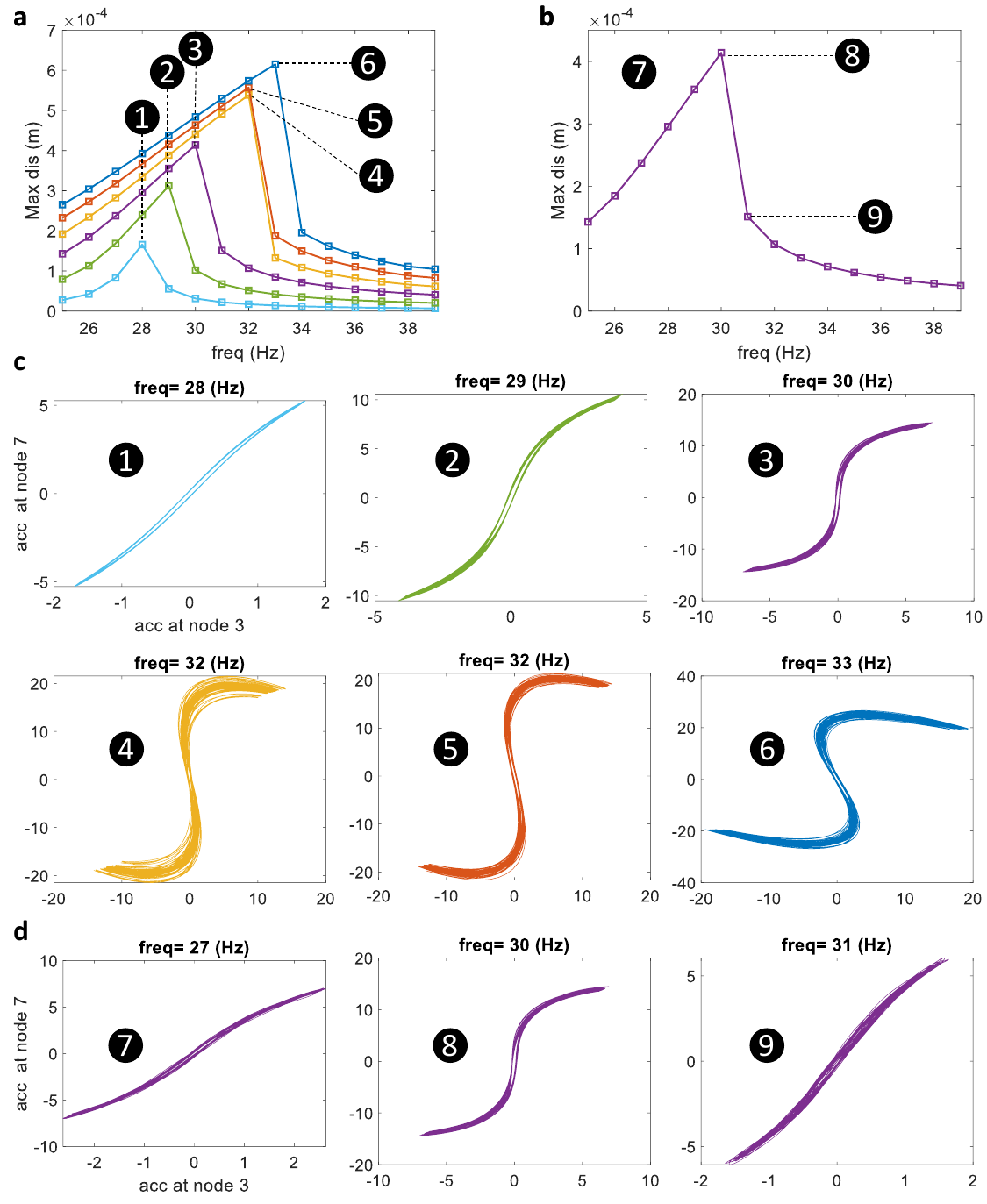}
	\caption{Nonlinear mode  isolation of the first NNM of a highly nonlinear beam. \textbf{a} First NNM for different energy levels. \textbf{b} NNM isolation at a specific energy level~(extracted from plot \textbf{a}) \textbf{c} Configuration plots for three different energy levels (marked in plot \textbf{a}). \textbf{d} Configuration plots of three different frequencies at a specific level of energy (as marked in plot \textbf{b})}
	\label{fig:num_first_mode}
\end{figure}

In this section, the non-linearity of the structure is increased and the first two isolated NNMs are obtained. The non-linearity coefficient $\mathbf{k_{nl}}$ is set to $15e9$. Fig. \ref{fig:num_first_mode} shows the plots for isolating the first NNM at different energy levels. Fig. \ref{fig:num_first_mode}(a) illustrates the maximum displacement of the main beam tip (node 7) at different frequencies surrounding the first natural frequency (27.7 Hz). As energy is increased, the first NNM deviates from the first LNM (light blue curve) and the magnitude of the first NNM increases as well since NNMs are energy-dependent phenomena. Fig. \ref{fig:num_first_mode}(c) illustrates the configuration plots of different NNMs shown in Fig. \ref{fig:num_first_mode}(a). As energy increases, the configuration of the isolated NNM becomes more twisted, indicating a higher level of nonlinearity. Fig. \ref{fig:num_first_mode}(d) depicts the three configuration plots for a fixed amount of energy (as seen in Fig. \ref{fig:num_first_mode}(b)): one below the NNM frequency, one at the NNM frequency, and one at a higher frequency than the NNM frequency. In comparison to the NNM frequency, configurations related to other frequencies are not as twisted. It should be noted that the frequency resolution used in numerical simulation is 1 Hz.
\begin{figure}[htbp]
	\centering
\includegraphics[width=1\textwidth]{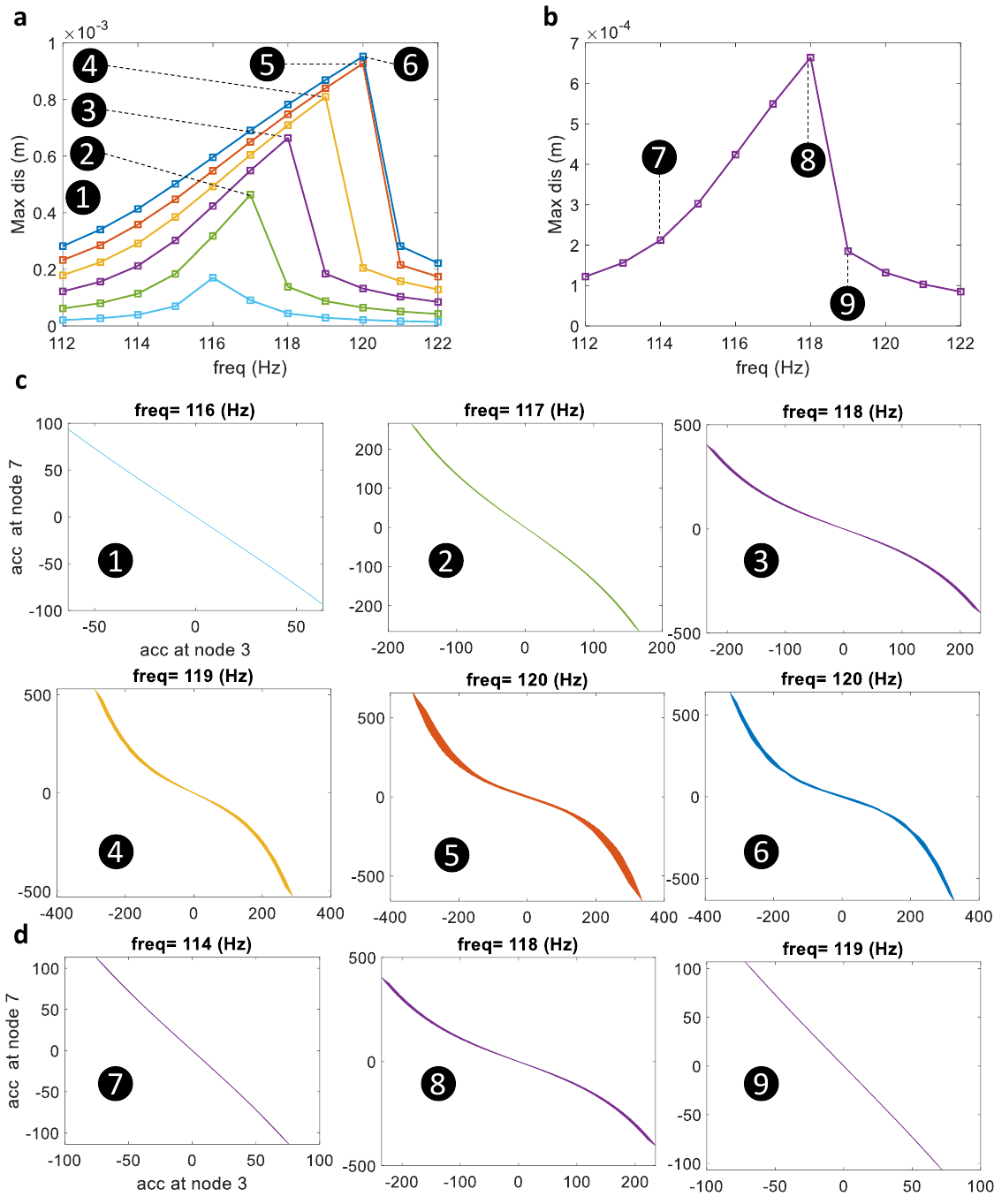}
	\caption{Nonlinear mode  isolation of the second NNM of a highly nonlinear beam. \textbf{a} Second NNM for different energy levels. \textbf{b} NNM isolation at a specific energy level~(extracted from plot \textbf{a}) \textbf{c} Configuration plots for three different energy levels (marked in plot \textbf{a}). \textbf{d} Configuration plots of three different frequencies at a specific level of energy (as marked in plot \textbf{b})}
	\label{fig:num_second_mode}
\end{figure}
\subsubsection{Numerical: Highly nonlinear structure: Second NNM isolations}
The process for extracting the second NNM is similar to that of the first NNM as shown in Fig. \ref{fig:num_second_mode}. By plotting the maximum displacement of the main beam tip at different frequencies~(Fig. \ref{fig:num_second_mode}(a)), the frequencies at which the second NNM occurs can be determined. As with the first NNM, the configuration plots corresponding to the second NNM show more twisted curves~(Fig. \ref{fig:num_second_mode}(c)), indicating a higher level of nonlinearity. Additionally, at a fixed energy level~(Fig. \ref{fig:num_second_mode}(b)), the configuration of the NNM frequency is more twisted than the configurations at other frequencies (lower and higher than the NNM frequency) as seen in Fig. \ref{fig:num_second_mode}(d).

\subsection{Experimental nonlinear beam}
\textcolor{black}{In this section, we present a data-driven approach for NNMs mode decomposition of an \textit{experimental} nonlinear beam using a physics-constraint deep neural network.} We describe an experimental setup for a nonlinear beam with similar characteristics to the numerical results previously presented. The structure is a weakly nonlinear beam and its dimensions and mechanical properties are identical to those reported for the numerical nonlinear beam~(as seen in Table. \ref{tab:2}).\par
\begin{figure}[htbp]
	\centering
\includegraphics[width=1\textwidth]{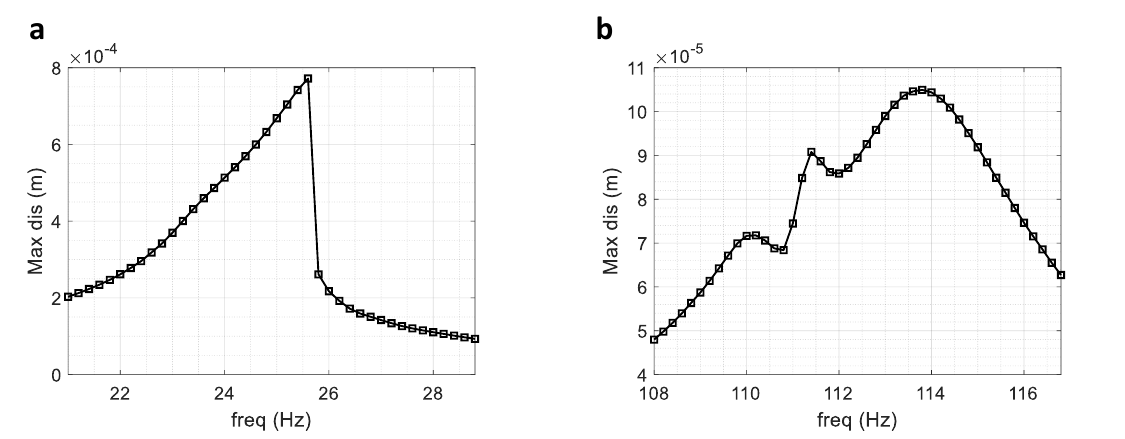}
	\caption{Sin-step frequency sweep method to extract the natural frequencies of the experimental nonlinear beam. \textbf{Left} First natural frequency. \textbf{Right} Second natural frequency}
	\label{fig:natural-freq}
\end{figure}

There are various methods for extracting natural frequencies, such as white noise excitation, impact testing, free vibration, etc~\cite{avitabile2001experimental}.  In this case, we use a low level of energy sin-step as an external force and create a plot similar to the one used before to isolate the NNM modes, which shows the maximum displacement at a specific point of the beam (in this case, the main beam tip) versus different frequencies. As shown in Fig. \ref{fig:natural-freq}, the first and second natural frequencies occur at 25.6 Hz and 113.8 Hz respectively, which are close to the numerical values. Next, we present the first and second NNMs extracted using the force appropriation method for the experimental weakly nonlinear beam.
\subsubsection{Experimental nonlinear beam: First and Second NNMs isolations}
In this section, the process of extracting the first and second NNM modes of an experimental nonlinear beam is outlined. This is achieved by using a step-sin external force induced by a shaker, as described in the numerical section. It's important to note that to maintain the structural integrity of the beam, a sufficient waiting period must be implemented between each frequency excitation. Additionally, due to the risk of permanent deformation, a high level of energy cannot be applied during the test. As a result, if the structure is not highly nonlinear, it may be difficult to observe nonlinearity at low energy levels.
\begin{figure}[htbp]
	\centering
\includegraphics[width=1\textwidth]{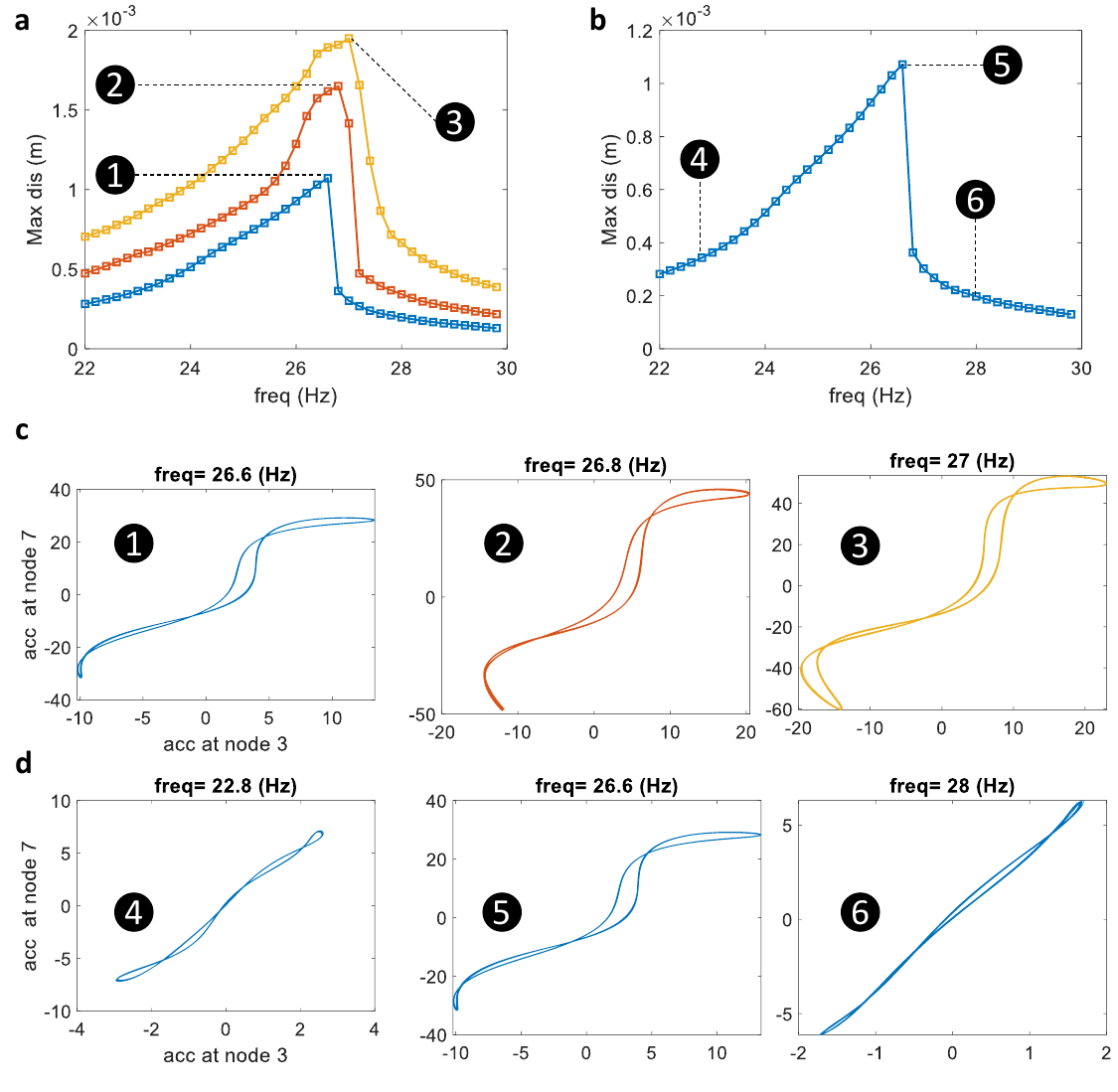}
	\caption{Nonlinear mode  isolation of the first NNM of the experimental nonlinear beam. \textbf{a} First NNM for different energy levels. \textbf{b} NNM isolation at a specific energy level~(extracted from plot \textbf{a}) \textbf{c} Configuration plots for three different energy levels (marked in plot \textbf{a}). \textbf{d} Configuration plots of three different frequencies at a specific level of energy (as marked in plot \textbf{b})}
	\label{fig:exp_first_mode}
\end{figure}

\begin{figure}[htbp]
	\centering
\includegraphics[width=1\textwidth]{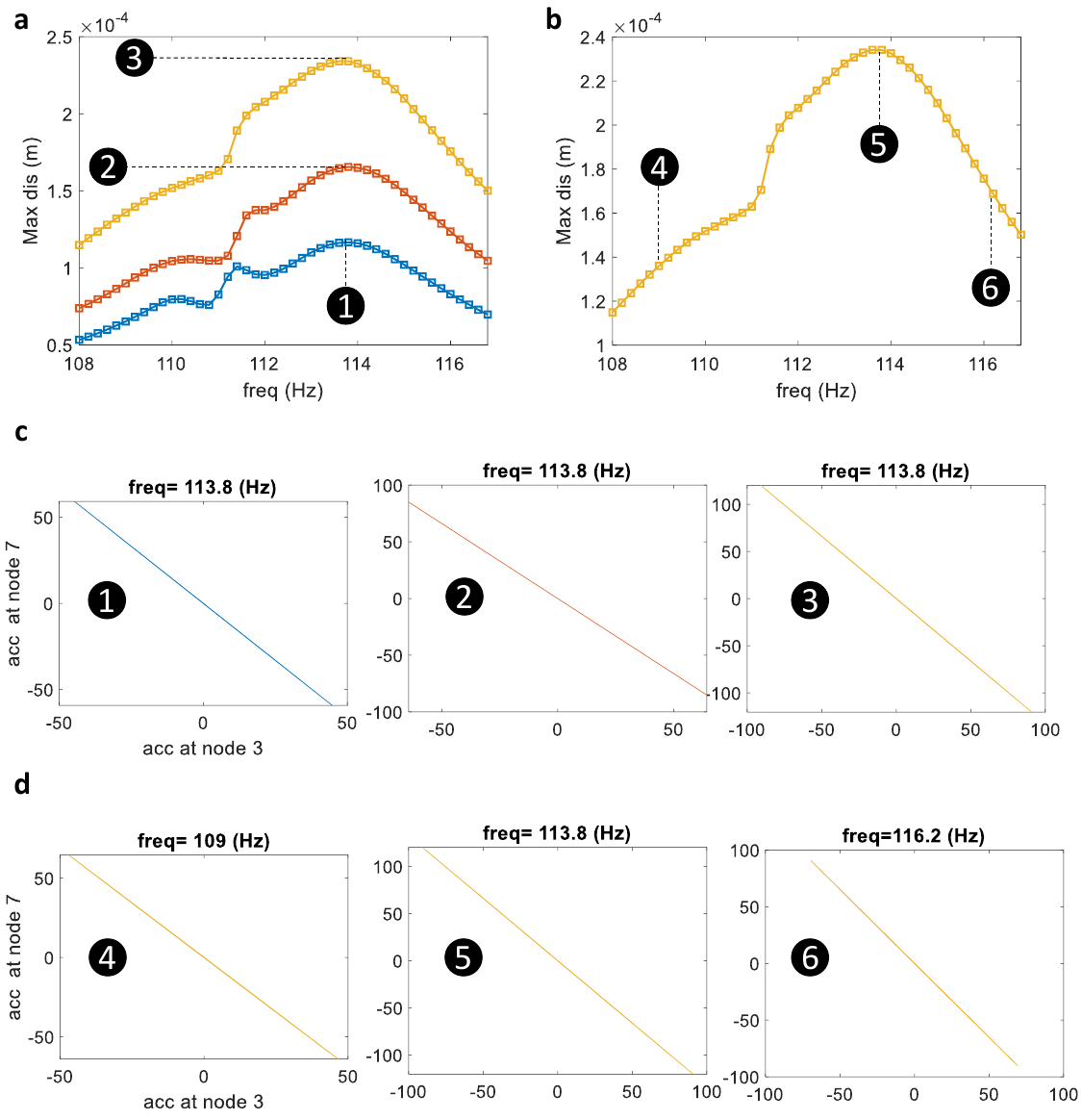}
	\caption{Nonlinear mode  isolation of the second NNM of the experimental nonlinear beam. \textbf{a} Second NNM for different energy levels. \textbf{b} NNM isolation at a specific energy level~(extracted from plot \textbf{a}) \textbf{c} Configuration plots for three different energy levels (marked in plot \textbf{a}). \textbf{d} Configuration plots of three different frequencies at a specific level of energy (as marked in plot \textbf{b})}
	\label{fig:exp_second_mode}
\end{figure}
Fig. \ref{fig:exp_first_mode} shows the results of isolating the first NNM mode of the experimental nonlinear beam. We can observe that as the energy level increases, the first NNM mode shifts from 26.6 to 27~(the frequency resolution used is 0.1 Hz). However, due to the beam's tendency to deform, we were unable to apply higher energy levels without affecting the structure's stiffness and other parameters. Nevertheless, the nonlinearity can be seen in Fig. \ref{fig:exp_first_mode}(c) and Fig. \ref{fig:exp_first_mode}(d) for different NNMs and frequencies at a fixed energy level, respectively.
For the second NNM mode, we followed the same procedure. However, as shown in Fig. \ref{fig:exp_second_mode}, the nonlinearity of this mode is weaker than that of the first NNM. The second NNM is isolated at three different energy levels in Fig. \ref{fig:exp_second_mode}(a), and the configuration plots in Fig. \ref{fig:exp_second_mode}(c) and Fig. \ref{fig:exp_second_mode}(d) for different energy levels and frequencies respectively, are almost flat, indicating a lack of significant nonlinearity in the second NNM.\par
It should be noted that the experimental beam used in this study is not identical to the one reported in previous literature~\cite{peeters2011dynamic}, nor is it the same as the numerical model discussed in the previous section in terms of nonlinearity. The experimental beam has higher damping compared to the numerical model, resulting in the second mode disappearing quickly and the nonlinearity being less pronounced. Additionally, the isolated modes in the experimental beam do not exhibit the same level of nonlinearity as seen in the numerical calculations due to limitations in applying high levels of energy to the experimental beam. This is because a large displacement during excitation can cause permanent deformation at the junction between the thin and main beams, potentially altering the structure's original stiffness and other parameters.
\subsubsection{Experimental nonlinear beam: NNMs Modes decomposition with NNMs-DNN}
\begin{figure}[htbp]
	\centering
\includegraphics[width=1.3\textwidth]{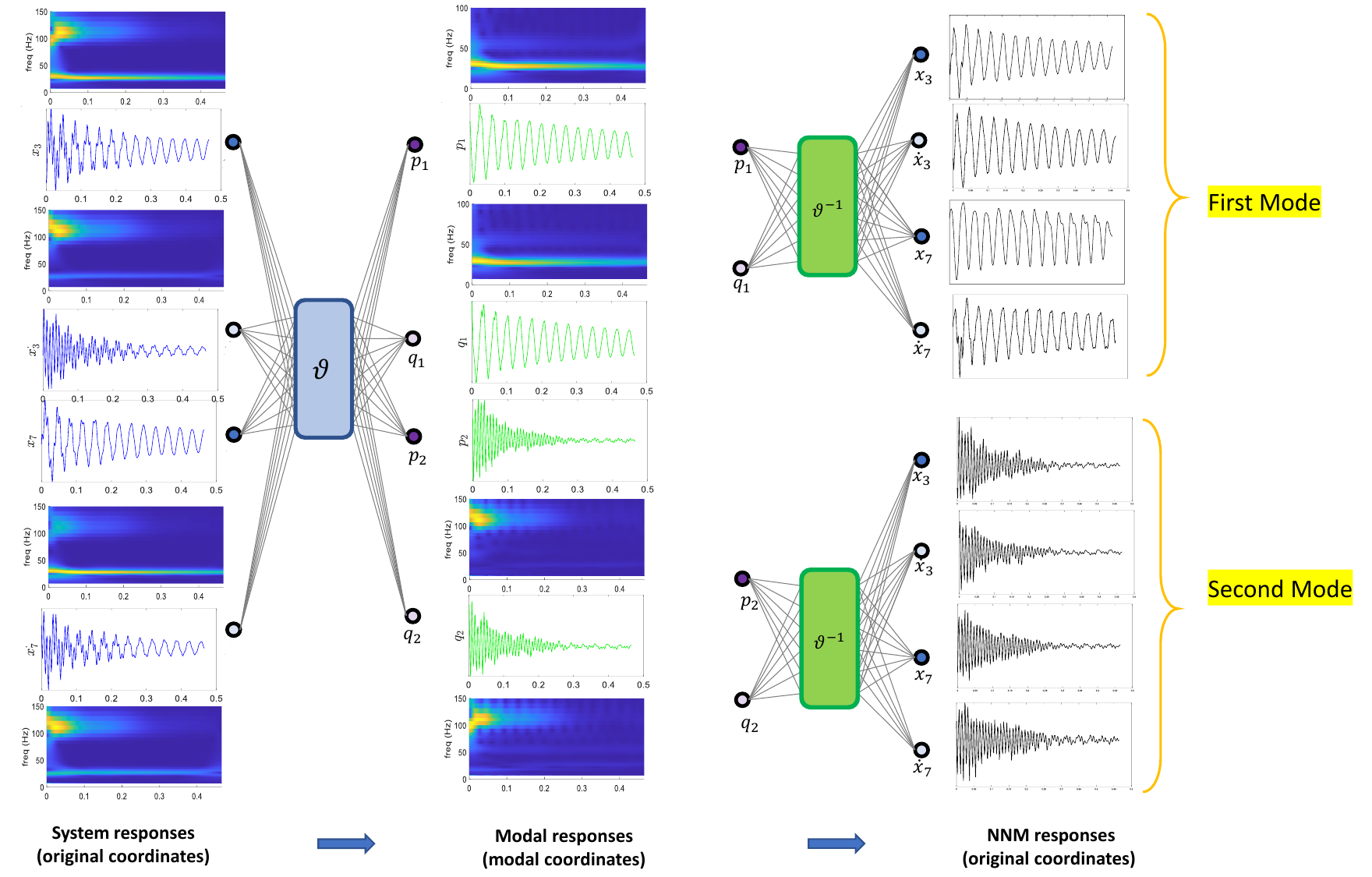}
	\caption{A demonstration of mode decomposition from the response of an experimental nonlinear beam using NNMs-DNN . The initial step involves utilizing the encoder to convert the input system response ${\mathrm{x}} = {\left[ {{x_3},{\dot x_3},{x_7},{\dot x_7}} \right]}$ into a modal space where each combination of modal displacement ${p_i}$ and modal velocity ${q_i}$ has a unique frequency. The next step involves separately using the decoder to convert each pair of modal responses (${p_i}$ and ${q_i}$) back to the original coordinates, which ultimately results in the output of the corresponding modal coordinates in the original space (in-phase and out-of-phase modal coordinates). \textcolor{black}{It should be noted that as we are looking for the first two NNMs, the input data is based on the free vibrations of two physical points~(nodes 3 and 7). }}
	\label{fig:mode-decomp}
\end{figure}

In this section, we aim to \textit{identify} the nonlinear modes of an experimental nonlinear beam using free vibration measurements~(from measurement data only). To achieve this, we used an impact hammer to excite the beam at location of node 2 and measured the accelerations at nodes 3 and 7 with accelerometers. \textcolor{black}{It is important to note that for the purpose of investigating the first two NNMs of the structure, free vibration measurements acquired from two physical points of the structure are sufficient.} We conducted 150 tests and recorded the free vibration for 0.5 seconds in each test. As shown in Fig. \ref{fig:mode-decomp}, the free vibration is characterized by two fundamental nonlinear frequencies. The first mode exhibits a significant degree of nonlinearity as the frequency decreases from 30 to 25.6 over time, whereas the second mode is less nonlinear, remaining around 113.8 Hz over time. 

To decompose the modes, the free vibration (mixed vibration) is fed into a neural network, and the separated modal coordinates are obtained in the latent space (output of the encoder). The latent spaces are forced to meet the requirements of NNMs through related loss functions. In the wavelet plots, it can be seen that each modal coordinate has a mono-frequency oscillation. To return to the original modal coordinates, only one pair of latent coordinates (one set of modal coordinates) is used and the other mode coordinates are frozen, then the decoder is used to restore the original coordinates. Fig. \ref{fig:mode-decomp} shows the in-phase (lower frequency) and out-of-phase (higher frequency) NNM modal coordinates obtained using our presented NNMs-DNN.
\subsubsection{Experimental nonlinear beam: Reconstruction and Prediction}
\begin{figure}[htbp]
	\centering
\includegraphics[width=1\textwidth]{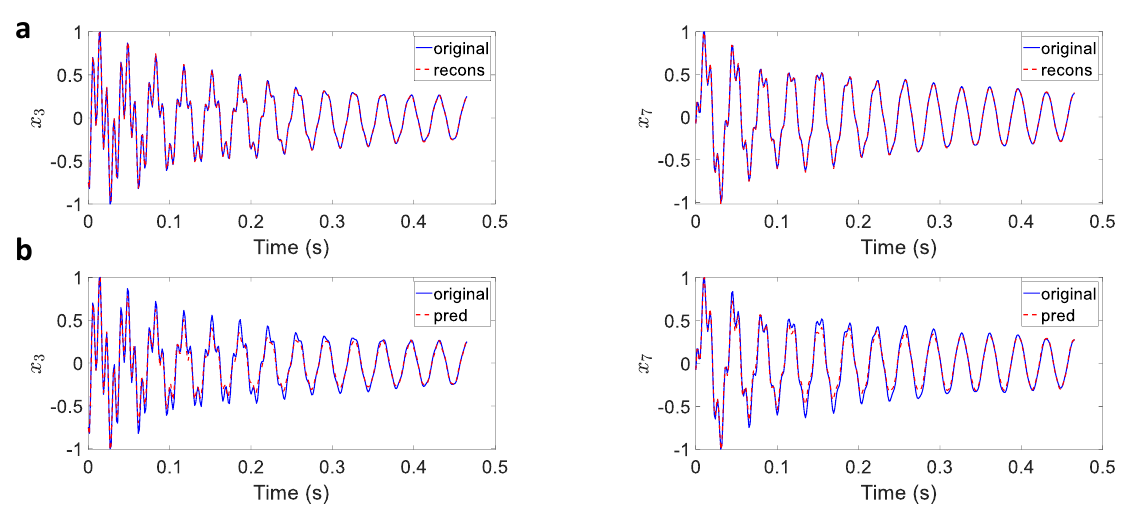}
	\caption{Reconstruction and prediction of free vibration responses of the experimental nonlinear beam  using our presented method. \textbf{a} Reconstructions corresponding to nodes 3 and 7. \textbf{b} Predictions corresponding to nodes 3 and 7. }
	\label{fig:rec-prd}
\end{figure}

\textcolor{black}{In this section, we showcase how our proposed framework can reconstruct and predict the free vibration of the structure, which are two important metrics used to evaluate the performance of our method. }The accuracy of the reconstruction is crucial to confirm that the latent space accurately represents the desired coordinates. Additionally, we can evaluate the prediction capability of the dynamics block to ensure it has a thorough understanding of the system's dynamics. Fig. \ref{fig:rec-prd} presents the results of our presented NNMs-DNN for nodes 3 and 7 for both reconstruction and prediction. The results from both reconstruction and prediction indicate that the network has effectively captured the dynamics and characteristics of the nonlinear beam. The dynamic block has a good grasp of the underlying physics of the structure, as it can predict 375 time steps consecutively.
\subsubsection{Training the network}
\begin{figure}[htbp]
	\centering
\includegraphics[width=1\textwidth]{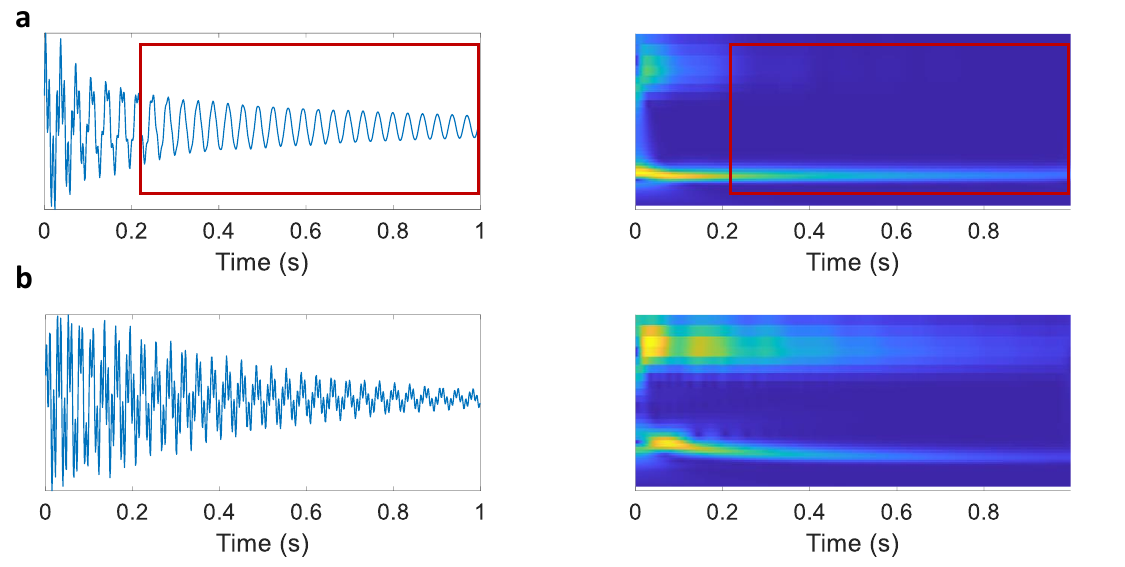}
	\caption{unbalanced experimental data \textbf{a} The red box indicates a lack of balance in the data, characterized by the presence of vibration at the first frequency and the absence of the second frequency. \textbf{b} numerical data: We simulated numerical data that is more balanced, where both frequencies contribute to the vibration of the nonlinear beam. Combining numerical and experimental data in the training phase can improve the ability of the DNN to perform mode decomposition effectively. }
	\label{fig:num_exp}
\end{figure}

For mode decomposition, we use an uncorrelated loss function which requires a reasonable length of the signal in terms of time. The second mode vanishes quickly over time as seen in Fig. \ref{fig:num_exp} (after 0.1 seconds, only the first mode remains), causing the free vibration to be unbalanced in terms of the existence of both modes. This makes it challenging for the network to learn how to decompose these modes. Using only the first portion of data where both modes exist (the first 0.1 seconds of free vibration) is not feasible since the frequency resolution, which is the inverse of the window length, would be 1/0.1 = 10 Hz. This means that the nonlinearity of the structure (the decreasing frequency over time) cannot be observed in wavelet plots as the range of frequency difference is around 4 Hz (from 30 Hz to 25.6 Hz). To overcome these issues, we created numerical datasets where the data is more balanced and the nonlinearity is more visible~(see Fig. \ref{fig:num_exp}). With this approach, the network can learn to distinguish frequency changes over time in the decomposition and handle unbalanced data. The figure displays both numerical and experimental data in terms of balance and nonlinearity.

\begin{table*}
	\caption{Geometrical \& Mechanical properties of beam}
	\label{tab:2}       
	\begin{tabular}{lp{2.5cm}p{2.5cm}p{2.5cm}p{2.5cm}p{2.5cm}}
		\hline\noalign{\smallskip}
	  & Length (m) & Width (m) & Thickness (m) & Young's modulus ($N/m^2$) & Density ($m/kg^3$)
   \\
\noalign{\smallskip}\hline\noalign{\smallskip}  \\
		
		Main beam & 0.7 & 0.01254 & 0.01254 & 1.605e11 & 8300   \\
		Thin beam  & 0.0406 & 0.0149 & 0.00051 &  1.605e11 & 8300
\\
\noalign{\smallskip}\hline

	\end{tabular}
\end{table*}
\section{Conclusion}
In this work, we studied the effectiveness of Nonlinear Normal Modes (NNMs) in characterizing nonlinear dynamics of structures, and a data-driven nonlinear modal analysis framework based on NNMs. We presented an experimental and numerical setup of a nonlinear beam for a comprehensive evaluation of the NNM framework and the data-driven nonlinear modal analysis framework. In the experiments, we presented a data-driven approach for mode isolation of both first and second NNM modes of an experimental nonlinear beam using step-sin external forces induced by a shaker. In addition, we applied the data-driven nonlinear modal analysis method (NNMs-embedded-DNN) to decompose the free vibration of the beam and separate the NNMs from response data only. Through the use of wavelet plots and reconstruction and prediction tests, we show that the NNMs-embeded-DNN successfully captured the modal dynamics and features of the nonlinear beam and understood the underlying physics of the structure. Overall, this study provides a useful experimental setup and method for extracting nonlinear modes of a nonlinear beam and demonstrates the effectiveness of the data-driven nonlinear modal analysis method (i.e., NNMs-embedded-DNN) for mode decomposition and dynamics prediction.

\textcolor{black}{The framework and experimental case study have potential limitations. Firstly, the training process is difficult due to the imbalanced experimental data of the nonlinear beam. Both modes are not equally incorporated, thus additional balanced data is needed to facilitate the training process. Additionally, it was not possible to subject the experimental beam to high levels of energy since this would result in the structure experiencing permanent high deflections, especially for the thin beam at the junctions with the main beam. The integration of energy-efficient and renewable resource-driven systems into smart building designs showcases how machine learning can enhance both environmental and operational performance in complex systems \cite{shafa6ranking}. These advancements highlight the impact of intelligent systems in overcoming conventional challenges, leading to greater efficiency and sustainability \cite{shafa6smart}. Lastly, in future works, more experimental case studies featuring higher nonlinearity and degrees of freedom could be used to assess the presented framework.}
\section*{Funding}
This research was partially funded by the Physics of Artificial Intelligence Program of U.S. Defense Advanced Research Projects Agency (DARPA) and the Michigan Technological University faculty startup fund.


 \bibliographystyle{elsarticle-num} 
 \bibliography{cas-refs}





\end{document}